\newif\ifneurips
\neuripsfalse %

\newcommand{\papertitle}{Learning Convex Optimization Models}

\documentclass[12pt]{article} %
\ifneurips
    \usepackage[nonatbib]{neurips_2020}
\fi

\usepackage{fullpage}
\usepackage{graphicx,subcaption,booktabs}

\title{\papertitle{}}
\author{Akshay Agrawal \and
Shane Barratt \and
Stephen Boyd\thanks{Authors listed in alphabetical order. Emails: \texttt{\{akshayka,sbarratt,boyd\}@stanford.edu}.}}

\newcommand{\inputs}{\mathcal X}
\newcommand{\outputs}{\mathcal Y}
\newcommand{\D}{\mathcal D}
\newcommand{\Dtrain}{\mathcal T}
\newcommand{\Dtest}{\mathcal V}

\usepackage[colorlinks,allcolors=blue,bookmarks=true,hypertexnames=true]{hyperref}
\usepackage{mathtools,graphicx,graphics,psfrag,amsmath,amsfonts,verbatim,xcolor,color}

\usepackage[
    natbib=true,
    citestyle=alphabetic,
    style=numeric,
    maxnames=3,
    minnames=3,
    maxbibnames=99,
    backend=bibtex,
    urldate=iso8601,
    firstinits=true,
    isbn=false,
    doi=false,
    date=year,
    sorting=nty
]{biblatex}
\DeclareFieldFormat[article,inbook,incollection,inproceedings,patent,thesis,
  unpublished]{citetitle}{#1}
\DeclareFieldFormat[article,inbook,incollection,inproceedings,patent,thesis,
  unpublished]{title}{#1}
\renewbibmacro{in:}{%
  \ifentrytype{article}{}{\printtext{\bibstring{in}\space}}}
\renewcommand*{\labelalphaothers}{\textsuperscript{+}}

\newcommand{\BEAS}{\begin{eqnarray}}
\newcommand{\EEAS}{\end{eqnarray}}
\newcommand{\BEA}{\begin{eqnarray}}
\newcommand{\EEA}{\end{eqnarray}}
\newcommand{\BEQ}{\begin{equation}}
\newcommand{\EEQ}{\end{equation}}
\newcommand{\BIT}{\begin{itemize}}
\newcommand{\EIT}{\end{itemize}}
\newcommand{\BNUM}{\begin{enumerate}}
\newcommand{\ENUM}{\end{enumerate}}

\newcommand{\BA}{\begin{array}}
\newcommand{\EA}{\end{array}}

\newcommand{\eg}{{\it e.g.}}
\newcommand{\ie}{{\it i.e.}}

\newcommand{\ones}{\mathbf 1}

\newcommand{\reals}{{\mbox{\bf R}}}

\newcommand{\symm}{{\mbox{\bf S}}}  %

\newcommand{\Expect}{\mathop{\bf E{}}}

\newcommand{\argmin}{\mathop{\rm argmin}}

\newcommand{\argmax}{\mathop{\rm argmax}}

\newcommand{\K}{\mathcal{K}}

\newcounter{algorithmctr}[section]
\renewcommand{\thealgorithmctr}{\thesection.\arabic{algorithmctr}}
   {\refstepcounter{algorithmctr}\begin{list}{}{%
       \setlength{\rightmargin}{0.03\linewidth}%
       \setlength{\leftmargin}{0.03\linewidth}}%
       \rmfamily\small
       \item[]{\setlength{\parskip}{0ex}\hrulefill\par%
        \nopagebreak{\bfseries\textsf{Algorithm \thealgorithmctr~}}}}%
   {{\setlength{\parskip}{-3ex}\nopagebreak\par\hrulefill} \end{list}}

\usepackage[T1]{fontenc}    %
\usepackage{microtype}      %

\begin{document}
\maketitle

\begin{abstract}
A convex optimization model predicts an output from an input by solving a
convex optimization problem.
The class of convex
optimization models is large, and includes as special cases many well-known models like
linear and logistic regression. We propose a heuristic
for learning the parameters in a convex optimization model from a
dataset of input-output pairs, using recently
developed methods for differentiating the solution of a convex optimization
problem with respect to its parameters.
\ifneurips
\else
We describe three general classes of convex optimization models,
maximum a posteriori (MAP) models,
utility maximization models, and agent models, and present a numerical experiment for each.
\fi
\end{abstract}

\section{Introduction}
\subsection{Convex optimization models}
We consider the problem of learning to predict outputs $y \in \outputs$ from
inputs $x \in \inputs$, given a set of input-output pairs $(x^i,
y^i)$, $i=1,\ldots,N$, with $(x^i,y^i) \in \inputs \times \outputs$.
We assume that $\outputs \subseteq \reals^m$
is a convex set, but make no assumptions on $\inputs$.
In this paper, we
specifically consider models $\phi : \inputs \to \outputs$
that predict the output $y$ by solving a convex optimization problem that
depends on the input $x$.  We call such models \emph{convex optimization models}.
While convex optimization has historically played a large role in \emph{fitting}
machine learning models, we emphasize that in this paper, we solve convex
optimization problems to perform \emph{inference}.

A convex optimization model has the form
\begin{equation}
  \phi(x;\theta)  = \argmin_{y\in \outputs} \, E(x, y; \theta),
  \label{eq:copt-model}
\end{equation}
where the objective function $E : \mathcal X \times \mathcal Y \to \reals \cup \{+\infty\}$
is convex in its second argument,
and $\theta$ is a parameter belonging to a set of 
allowable parameters $\Theta$.
The objective function $E$ is the model's \emph{energy function},
and the quantity $E(x, y; \theta)$ is
the energy of $y$ given $x$; the energy $E(x, y; \theta)$ can depend arbitrarily on $x$
and $\theta$, as long as it is convex in $y$.
Infinite values of $E$ encode additional constraints on the
prediction, since $E(x, y; \theta) = +\infty$ implies $\phi(x;\theta) \neq y$.
Evaluating a convex optimization model
at $x$ corresponds to finding an output $y \in \outputs$ of minimum energy.
The function $\phi$ is in general
set-valued, since the convex optimization problem in \eqref{eq:copt-model} may
have zero, one, or many solutions. Throughout this paper, we only consider
the case where the argmin exists and is unique.

Convex optimization models
are particularly well-suited for problems in which
the outputs $y \in \outputs$ are known to have structure. For example,
if the
outputs are probability mass functions, we can take $\outputs$ to be the
probability simplex; if they are sorted
vectors, we can take $\outputs$ to be the monotone cone; or if they are
covariance matrices, we can take $\outputs$ to be the set of symmetric positive
semidefinite matrices. In all cases, convex optimization models provide an
efficient way of searching over a structured set to produce predictions
satisfying known priors.

Because convex optimization models can depend arbitrarily on $x$ and $\theta$,
they are quite general. We will see that they include familiar 
models for regression and classification,
such as linear and logistic regression, as specific instances.
In the basic examples of linear and logistic regression, the corresponding
convex optimization models have analytical solutions.
But in most cases, convex optimization models must be evaluated by
a numerical algorithm.

\ifneurips
\else
Learning a parametric model requires tuning the parameters to make good
predictions on $\mathcal D$ and ultimately on held-out input-output pairs.
\fi
In this paper, we present a gradient method for learning the parameters in
a convex optimization model; this learning problem is in general non-convex,
since the solution map of a convex optimization model is a complicated function.
Our method uses the fact that the solution map is often differentiable,
and its derivative can be computed efficiently, without differentiating through
each step of the numerical solver \citep{agrawal2019differentiable, diffcp2019,
amos2019differentiable, diffcp2019, busseti2018solution}.
\ifneurips
\else
\paragraph{Outline.}
Our learning method is presented in \S\ref{s:learn} for the general
case.
In the following three sections, we describe general classes of
convex optimization models with particular forms or interpretations.
In \S\ref{s:map}, we interpret convex optimization models as solving a
maximum a posteriori (MAP) inference task,
and we give examples of these MAP
models in regression, classification, and graphical models. In
\S\ref{sec:utility_max}, we show how convex optimization models can be used to
model utility-maximizing processes.  
In \S\ref{s:stoch-control}, we give
examples of modeling agents using the framework of stochastic control. 
In \S\ref{s:num-exp}, we present numerical experiments of learning convex
optimization models for several prediction tasks.
\fi

\subsection{Related work}\label{s:rel-work}

\paragraph{Structured prediction.}
Structured prediction refers to supervised learning problems
where the output has known structure \cite{bakir2007predicting}.
A common approach to structured prediction is energy-based models,
which associate a scalar energy to each output,
and select a value of the output that minimizes the energy,
subject to constraints on the output \cite{lecun2006tutorial}.
Most energy-based learning methods are learned by reducing the energy for input-output pairs in the training set and increasing it
for other pairs \cite{taskar2004max,taskar2005learning,tsochantaridis2005large,chopra2005learning}.
More recently, the authors of \cite{belanger2016structured,belanger2017end}
proposed a method for end-to-end learning of energy networks by unrolled optimization.
Indeed, a convex
optimization model can be viewed as a form of energy-based learning
where the energy function is convex in the output.
For example, input-convex neural networks
(ICNNs) \cite{amos2017input} can be viewed as a convex optimization model
where the energy function is an ICNN. We also note that several authors have
proposed using structured prediction methods as the final layer of a deep
neural network \cite{peng2009conditional,zheng2015conditional,chen2015learning}; of particular note is \cite{geng2020coercing},
in which the authors used a second-order cone program (SOCP) as their final layer.

\paragraph{Inverse optimization.}
Inverse optimization refers to the problem of recovering the structure or parameters
of an optimization problem, given solutions to it \cite{ahuja2001inverse, heuberger2004inverse}.
In general, inverse optimization is very difficult. One special case where it is tractable is when the optimization
problem is a linear program and the loss function is convex in the
parameters \cite{ahuja2001inverse}, and another is when the optimization
problem is convex and the parameters enter in a certain way \cite{chatalbashevinverse, keshavarz2011imputing}.
This paper can be viewed as a heuristic method for inverse optimization for general convex optimization problems.

\paragraph{Differentiable optimization.}
There has been significant recent
interest in differentiating the solution maps of optimization problems;
these differentiable solution maps are sometimes called \emph{optimization layers}.
The paper \cite{amos2017optnet} showed how quadratic programs can be embedded
as optimization layers in machine learning pipelines, by implicitly
differentiating the KKT conditions (as in the early works
\cite{fiacco1968nonlinear, fiacco1976sensitivity}). Recently,
\cite{diffcp2019, amos2019differentiable} showed
how to efficiently differentiate through convex cone programs by applying the
implicit function theorem to a residual map introduced in
\cite{busseti2018solution}, and \cite{agrawal2019differentiable} showed how
to differentiate through convex optimization problems by an automatable
reduction to convex cone programs; our method for learning convex
optimization models builds on this recent work. Optimization layers have
been used in many applications, including control \cite{
amos2018differentiable, de2018end, barratt2019fitting,
cocp2020}, game-playing \cite{ling2018game, ling2019large},
computer graphics \cite{geng2020coercing}, combinatorial tasks
\cite{vlastelica2020differentiation, rolinek2020optimizing, rolinek2020deep, berthet2020learning},
automatic repair of optimization problems \cite{barratt2020automatic}, and data fitting more generally \cite{amos2017input, barratt2018optimizing,
barratt2019least, amos2019differentiablecross}.
Differentiable optimization for nonconvex problems is often performed
numerically by differentiating each individual step of a numerical solver
\cite{domke2012generic,maclaurin2015gradient,diamond2017unrolled,finn2018learning},
although sometimes it is done implicitly; see, \eg,
\cite{amos2018differentiable,lorraine2019optimizing,agrawal2020differentiating}.

\paragraph{Bilevel optimization.} The task of minimizing the training error
of a convex optimization model can be interpreted as a \emph{bilevel}
optimization problem, \ie, an optimization problem in which some of the
variables are constrained to be optimal for another optimization problem
\cite{colson2007overview}. In our case, the optimization problem is to minimize
the model's training error, subject to the constraint that the predicted output
is the solution to a convex optimization problem.

\section{Learning convex optimization models}
\label{s:learn}
In this section we describe a general
method for learning the parameter $\theta$ in a convex optimization model,
given a data set consisting of input-output pairs $(x^1,y^1), \ldots,
(x^N,y^N) \in \inputs \times \outputs$.
We let $\hat y^i = \phi(x^i;\theta)$ denote the prediction of $y^i$
based on $x^i$, for $i=1, \ldots, N$.
These predictions depend on $\theta$, but we suppress this dependency to
lighten the notation.

\subsection{Learning problem}
The fidelity of a convex optimization model's predictions
is measured by a loss function
$L : \outputs \times \outputs \to \reals$.
The value $L(\hat y^i, y^i)$ is the
loss for the $i$th data point; the lower the loss,
the better the prediction.
Through $\hat y^i$, this depends on the parameter $\theta$.

Our ultimate goal is to construct a model that generalizes, \ie,
makes accurate predictions for input-output pairs not present in $\D$.
To this end, we first partition the data pair indices
into two sets, a training set $\Dtrain \subset \{1, \ldots, N\}$
and a validation set $\Dtest = \{1, \ldots, N\} \setminus \Dtrain$.
We define the average training loss as
\ifneurips
\else
\[
\mathcal L (\theta) = \frac{1}{|\Dtrain|} \sum_{i \in \Dtrain}
L(\hat y^i, y^i).
\]
\fi
We fit the model by choosing $\theta$ to minimize the
average training loss plus a regularizer $R : \Theta \to \reals \cup
\{\infty\}$,
\ie, solving the optimization problem
\begin{equation}\label{e-theta-fit-prob}
\begin{array}{ll}
\mbox{minimize} & \mathcal L (\theta) + R(\theta),
\end{array}
\end{equation}
with variable $\theta$.
The regularizer measures how compatible $\theta$ is with prior knowledge,
and we assume that $R(\theta)=\infty$ for $\theta \not\in \Theta$,
\ie, the regularizer encodes the constraint $\theta \in \Theta$.
We describe below a gradient-based method to (approximately)
solve the problem~(\ref{e-theta-fit-prob}).

We can check how well a convex optimization model generalizes
by computing its average loss on the validation set,
\ifneurips
\else
\[
\mathcal{L}^\mathrm{val}(\theta) = \frac{1}{|\Dtest|}
\sum_{ i \in \Dtest} L(\hat y^i, y^i).
\]
\fi
In some cases, the model or learning procedure
depends on parameters other than $\theta$, called
hyper-parameters. It is common to learn multiple models
over a grid of
\ifneurips
\else
hyper-parameter
\fi
values and use the model with the
lowest validation  loss.

\subsection{A gradient-based learning method}
In general, $\mathcal L$ is not convex, so we must
resort to an approximate or heuristic method for learning the parameters. One
could consider zeroth-order methods, \eg, evolutionary strategies
\cite{hansen2001completely}, Bayesian optimization \cite{movckus1975bayesian},
or random search \cite{solis1981minimization}. Instead, we use a first-order
method, taking advantage of the fact that the convex optimization model is
often differentiable in the parameter $\theta$.

\paragraph{Differentiation.}
The output of a non-pathological
convex optimization model is an implicit function of the
input $x$ and the parameter $\theta$. When some regularity conditions are
satisfied, this implicit function is differentiable, and its derivative with
respect to $\theta$
can often be computed in less time than is needed to compute the solution \cite{diffcp2019}. One generic way
of differentiating through convex optimization problems involves a reduction to
an equivalent convex cone program, and implicit differentiation of a
residual map of the cone program \cite{diffcp2019}; this is the method we use in this
paper. For readers interested in more details on the
derivative computation, we suggest \cite{busseti2018solution,
diffcp2019, agrawal2019differentiable}.
In our experience, it is unnecessary to check regularity conditions, since we
and others have empirically observed that the derivative computation in
\cite{diffcp2019} usually provides useful first-order information
in the rare cases when the solution map is not
differentiable at the current iterate
\cite{agrawal2019differentiable, cocp2020}. In this sense, convex optimization
models are similar to other kinds of machine learning models, such as neural
networks, which can be trained using gradient descent despite only being
differentiable almost everywhere.

\paragraph{Learning method.}
We propose a proximal stochastic gradient method.
The method is iterative, starting with an initial parameter $\theta^1$.
The first step in iteration $k$ is to choose a batch of (training)
data denoted $\mathcal B^k\subset \mathcal T$.
There are many ways to do this, \eg, by cycling through the training
set or by selecting indices in $\mathcal T$ at random.
The next step is to compute the gradient of the loss
averaged over the batch,
\ifneurips
\else
\[
g^k = \frac{1}{|\mathcal B^k|}\sum_{i \in \mathcal B^k}
\nabla_\theta L(\hat y^i,y^i).
\]
\fi
This step requires applying the chain rule for differentiation
to $|\mathcal B^k|$ compositions
of the convex optimization model (discussed above) and the loss function.
The final step is to update $\theta$ by first taking a step in the
negative gradient direction, and then applying the proximal operator of $R$,
\[
\theta^{k+1} = \mathbf{prox}_{t^kR}(\theta^k - t^k \nabla g^k) =
 \argmin_{\theta \in \Theta} \,  \left( R(\theta) + 
\frac{1}{2t^k}\|\theta - (\theta^k - t^k \nabla g^k)\|_2^2 \right),
\]
where $t^k>0$ is a step size.
We assume that the proximal operator of $R$ is single-valued
and easy to evaluate.
When $R(\theta)$ is the $\{0,\infty\}$ indicator function of
$\Theta$, this method reduces
to the standard projected stochastic gradient method,
\ifneurips
\else
\[
\theta^{k+1} = \Pi_\Theta(\theta^k - t^k \nabla g^k),
\]
\fi
where $\Pi_\Theta$ is the Euclidean projection operator onto $\Theta$.
There are many ways to select the step sizes $t^k$; see,
\eg, \cite{nocedal2006numerical,beck2009gradient,bottou2010large}.

\section{MAP models}\label{s:map}

Let the inputs $x \in \inputs$ and outputs $y \in
\outputs$ be random vectors, and suppose that the conditional distribution of $y$
given $x$ has a log-concave density $p$, parametrized by $\theta$. The energy
function
\[
  E(x, y; \theta) = -\log p(y \mid x; \theta)
\]
yields a \emph{maximum a posteriori} (MAP) model:
$\hat y = \phi(x; \theta)$
is the MAP estimate of the random vector $y$, given $x$ \cite[\S 1.2.5]{bishop2006pattern}.
Conversely, any convex
optimization model can be interpreted as a MAP model, by identifying
the density of $y$ given $x$ with
\ifneurips
\else
an exponential transformation
\fi
of the negative energy
\[
  p(y \mid x; \theta) = \frac{1}{Z(x; \theta)}\exp(-E(x, y; \theta)),
\]
where
\[
  Z(x; \theta) = \int_{y \in \outputs} \exp(-E(x, y; \theta)) \; dy
\]
is the normalizing constant or partition function.
Crucially, evaluating a MAP model does not require computing $Z(x; \theta)$
since it does not depend on $y$; \ie, MAP models can be used even when
the partition function is computationally intractable, as is often the case \cite[\S18]{goodfellow2016deep}.

\subsection{Regression}
\label{sec:regression}
Several basic regression models can be described as MAP models, with
\[
  p(y \mid x) \propto \exp(-f(\theta^T x - y)),
\]
where $x \in \inputs = \reals^n$, $y \in \outputs = \reals^m$, $\theta \in
\reals^{n \times m}$ is the parameter and $f : \reals^{m} \to \reals$ is a
convex penalty function. (The expression $\theta^T x$ can be
replaced with a more complex function, such as a neural network, since convex
optimization models can depend arbitrarily on $x$ and $\theta$; we focus on the
linear case for simplicity.)
If the penalty $f$ is minimized at 0,
then the MAP model is the linear predictor $\phi(x; \theta) = \theta^Tx$.
In this case, fitting the MAP model with a mean-squared loss $L(\hat y, y) = \|\hat y - y\|_2^2$
is equivalent to fitting a linear regression model; fitting it with an
$\ell_1$ loss
is equivalent to $\ell_1$ regression; and fitting it with the Huber loss
 \cite[\S 6.1]{boyd2004convex} yields robust Huber regression.

These very basic examples can be made more interesting by constraining the outputs $y$ to lie in a convex subset
$C$ of $\outputs$, using a density of the form
\ifneurips
\else
\[
  p(y \mid x) \propto
  \begin{cases}
    \exp(-f(\theta^T x - y))  & y \in C \\
    0 & \text{otherwise}.
  \end{cases}
\]
\fi
Because the output is constrained, different choices of the penalty
function $f$ yield different MAP models. When the penalty function $f$ is the squared Euclidean norm, $f(u)=\|u\|_2^2$,
the MAP estimate $\hat y = \phi(x; \theta)$ is the Euclidean projection of $\theta^T x$
onto $C$. Other penalty functions, like the $\ell_1$ norm $f(u) = \|u\|_1$
or the Huber function \cite[\S 6.1]{boyd2004convex}
yield interesting non-trivial regression models. We present some examples
of the constraint set $C$ below.

\paragraph{Nonnegative regression.} Taking $C = \reals^{m}_{+}$ (the set of
nonnegative $m$-vectors) yields a MAP model for nonnegative regression, \ie,
the MAP estimates in this model are guaranteed to be nonnegative.

\paragraph{Monotonic output regression.}
When $C$ is the monotone cone, \ie, the set of ordered vectors
\ifneurips
\else
\[
  C = \{y \in \reals^{m} \mid y_1 \leq y_2 \leq \cdots \leq y_m\},
\]
\fi
the MAP estimates in the regression model are guaranteed to be sorted
in ascending order.
When $f$ is the Euclidean norm, the MAP estimate is the projection of
$\theta^T x$ onto the monotone cone, and evaluating it
requires solving a convex quadratic program (QP); in this special case, once
$\theta^Tx$ has been computed (which takes $O(mn)$ time), evaluating the convex optimization model is equivalent to monotonic or isotonic regression \cite{barlow1972isotonic}, which takes $O(m)$ time \cite{best1990active},
meaning it has the same complexity as the standard linear regression model.

\ifneurips
\else
We note the distinction between traditional isotonic regression
\cite{barlow1972isotonic} and a convex optimization model with monotone
constraint.  In isotonic regression, we seek a single vector with
nondecreasing components.  In a convex optimization model with
a monotone constraint, we seek a model that maps $x \in \inputs$ to
a prediction $\hat y$ that always has nondecreasing components.
\fi

\subsection{Classification}
\label{sec:classification}
In (probabilistic) classification tasks,
the outputs are vectors in the probability simplex, \ie,
\ifneurips
\else
\[
  \mathcal{Y} = \Delta^{m-1} = \{y \in \reals^{m} \mid \ones^T y = 1, ~ y \geq 0\}.
\]
\fi
The output $y$ can be interpreted as
a probability distribution over $\{1,\ldots,m\}$ associated with an input $x \in \inputs = \reals^n$. The MAP estimate
$\hat y = \phi(x; \theta)$ is therefore the most likely distribution associated with
$x$, under a particular density $p(y \mid x; \theta)$.
This includes as a special case the familiar setting in which each output is a
label, \eg, a number in $\{1, \ldots, m\}$, since the label $k$ can be
represented by a vector $y$ such that $y_k = 1$ and $y_i = 0$ for $i \neq k$.

As a simple first example, consider the MAP model with density
\[
  p(y \mid x; \theta) \propto
  \begin{cases}
    \exp(x^T \theta y + H(y)) & y \in \Delta^{m-1} \\
    0 & \text{otherwise},
  \end{cases}
\]
where $\theta \in \reals^{n \times m}$ and $H(y)=-\sum_{i=1}^{m} y_i \log y_i$ is the entropy function.
The resulting convex optimization model is
just the softmax of $\theta^T x$, \ie, $\phi(x;\theta) = \exp(\theta^T x) / \ones^T \exp(\theta^Tx)$,
where the exponentiation and the division are meant elementwise. (This fact is
readily verified via the KKT conditions of the convex optimization model
\citep[\S2.4.4]{amos2019differentiable}).

Since the outputs are probability distributions, a natural loss function is the
KL-divergence from the true output $y$ to the prediction $\hat y = \phi(x;
\theta)$, \ie,
\[
L(\hat y, y) = \sum_{i=1}^{m} y_i \log (y_i/\hat y_i) = \sum_{i=1}^{m} y_i \log y_i - y_i \log \hat y_i.
\]
Discarding the constant terms $y_i \log y_i$, which do not affect learning,
recovers the commonly used cross-entropy loss \cite[\S2.6]{hastie2009elements}. Using this loss function
with the softmax model recovers multinomial logistic regression \cite[\S4.4]{hastie2009elements}.
This model can be made more interesting by simple extensions.

\paragraph{Constrained logistic regression.} We can readily add constraints
on the distribution $\hat y$.  As a simple example,
a box-constrained logistic regression model has the form
\[
  \begin{array}{llll}
   \phi(x;\theta) &=& \mbox{argmin}_y & -x^T \theta y - H(y) \\
                  && \mbox{subject to} & y \in C,
 \end{array}
\]
where $C$ is a convex subset of $\Delta^{m-1}$. There are many interesting
constraints we can impose on the distribution $y$. As a simple example, the
constraint set
\ifneurips
\else
\[
  C = \{y \in \Delta^{m-1} \mid \alpha \leq y \leq \beta\},
\]
\fi
where $\alpha, \beta \in \reals^{m}$ are vectors and the
the inequalities are meant
elementwise can be used to require
that $\hat y$ have heavy tails, by making the leading and trailing components of
$\alpha$ large, or thin tails, by making the leading and trailing components of
$\beta$ small. Another simple example is to specify the expected value of an
arbitrary function on $\{1, \ldots, m\}$ under $\hat y$, which is a simple
linear equality constraint on $\hat y$. More generally, any affine equality
constraints and convex inequality constraints on $\hat y$ may be imposed;
these include constraints on the quantiles of the random
variable associated with $y$, lower bounds on its variance, and inequality
constraints on conditional probability distributions.

\paragraph{Piecewise-constant logistic regression.}
A piecewise-constant logistic regression model has the form
\[
  \begin{array}{llll}
   \phi(x;\theta) &=& \mbox{argmin}_y & -x^T \theta y - H(y) + \lambda \sum_{i=1}^{m-1} |y_{i+1}-y_i| \\
                  && \mbox{subject to} & y \in \mathcal Y,
 \end{array}
\]
where the parameter is $\theta$ and $\lambda > 0$ is a (hyper-)parameter.
To the standard energy we add a total variation term that
encourages $y$ to have few ``jumps'', \ie, few indices $i$ such that $y_i \neq y_{i+1}$,
$i = 1, \ldots, m-1$ \cite[\S7.4]{kim2009ell_1}. The larger the
hyper-parameter $\lambda$ is, the fewer jumps it will have (typically).

\subsection{Graphical models}\label{s:graph}
A Markov random field (MRF) is an undirected graphical model
that describes the joint distribution of a set of random variables, which
are represented by the nodes in the graph. An MRF associates parametrized
potential functions to cliques of nodes, and the joint distribution it describes is
proportional to the product of these potential functions. MRFs are commonly
used for structured prediction, but learning their parameters is in
general difficult \cite[\S8.3]{bishop2006pattern}. When the potential
functions are log-concave, however, we can fit the parameters using the methods
described in this paper.

Suppose we are given an MRF describing the joint distribution of the random vectors
$x$ and $y$. Let $z = (x, y) \in \reals^{n+m}$, and let $c_1$, $c_2$, \ldots, $c_p$ denote the
indices of the graph cliques; we write $z_{c_k}$ to denote the components of $z$
in clique $c_k$. For example, if $c_k = (1, 4, 5)$, then $z_{c_k} = (z_1,
z_4, z_5)$. Suppose the MRF has a Boltzmann distribution, meaning
\[
  p(y \mid x; \theta) \propto \exp(-(E_1(z_{c_1}) + E_2(z_{c_2}) + \cdots + E_p(z_{c_p}))).
\]
Here, $\exp(-E_k(z_{c_k}))$ are the potential functions, and $E_k$ is
a local energy function, parametrized by $\theta$, for the clique
$k$. As long as the functions $E_1, \ldots, E_p$ are convex, the corresponding
MAP model
\[
  \phi(x; \theta) = \argmax_{y \in \outputs} \log p(y \mid x; \theta)
\]
is a convex optimization model. In this case, given a dataset of input-output
pairs $(x, y)$, we can fit the parameter $\theta$ without evaluating or
differentiating through the partition function.

\paragraph{Quadratic MRFs.}
Consider an MRF in which the variables $x$ and $y$ lie in convex sets (such
as slabs, or all of $\reals^{n}$ or $\reals^{m}$). Suppose the MRF has ${{n+m}\choose{2}} +
n+m$ pairwise cliques of the form $\{z_i, z_j\}$ ($1 \leq i \leq j \leq n + m$), and
a Boltzmann distribution with local energy functions
\[
E_{(i, j)}(z_i, z_j) = \theta_{ij} z_iz_j, \quad 1 \leq i \leq j \leq n + m,
\]
where $\theta \in \Theta = \symm^{n+m}_{+}$ is the parameter ($\symm^{n+m}_{+}$
is the set of positive semidefinite matrices). The MAP inference task for
this MRF is a convex optimization model, of the form
\[
  \phi(x;\theta) = \argmax_{y \in \outputs} \, -z^T\theta z = \argmin_{y \in \outputs} \, z^T \theta z.
\]
MRFs with a similar clique structure have been proposed for various signal
and image denoising tasks. We give a numerical example of fitting a quadratic
MAP model of an MRF in \S\ref{s:num-exp}.
\ifneurips
\else

\fi
We emphasize that the dependence on $x$ can be arbitrary; \eg, if the
energy function were
\ifneurips
\else
\[
  E(x, y; \theta) = (f(x), y)^T \theta (f(x), y)),
\]
\fi
where $f$ were a neural network, the MAP model would remain convex.

%
%
%

\ifneurips
\else
\section{Utility maximization models}
\label{sec:utility_max}
We now consider the case where the output $y$
is interpreted as a decision, and the input $x$ is a context or feature vector
that affects the decision.
We assume that the decision $y$ is chosen to maximize
some given parametrized utility function
\[
U:\mathcal X\times\mathcal Y\to\reals\cup\{-\infty\},
\]
where $U(x,y;\theta)$ is the utility of choosing a decision $y$ given
the context $x$ and the parameters $\theta$,
and is concave in $y$.
Infinite values of $U$ are used to constrain the decision $y$.
(In most cases the utility function $U$ is monotone increasing in $y$,
but we do not need this property.)
The energy function in a utility maximization model is simply
the negative utility,
\[
E(x,y;\theta) = -U(x,y;\theta).
\]
The resulting convex optimization model $\phi(x;\theta)$
gives a maximum utility decision in the context $x$.
The same losses used for regression (see \S\ref{sec:regression}) and classification (see \S\ref{sec:classification}) can be used for utility maximization.
The context $x$ might include, for example, a total budget on
the decision $y$, prices that affect the decision, or
availabilities that affect the decision.

\paragraph{Resource allocation.}
A standard example of utility maximization is resource allocation.
In the simplest case, this involves allocating a single, finite resource
across $m$ agents or tasks. The decision $y\in\reals_+^m$ gives the allocation
across those tasks, where $y_i$ is the resource allocated to task $i$;
because the resource is finite, the allocation must satisfy $\ones^T y \leq B$,
where $B \in \reals_{+}$ is a nonnegative budget.
The context $x$ contains the budget $B$,
and possibly other important parameters such as limits
on allocations to the tasks.
When the input $x$ is just the budget, the utility has the form
\[
U(x,y;\theta) = \begin{cases}
U(y;\theta) & y \geq 0, \quad \ones^Ty \leq B, \\
-\infty & \text{otherwise},
\end{cases}
\]
where $U(y;\theta)$ is some parametrized concave utility function,
describing the utility of an allocation. In this simple case, $\phi(x;\theta)$
gives the maximum utility allocation that satisfies the budget constraint.

The input $x$ is not limited to just the budget; it can also contain additional
context that affects or constrain the decision.
One important case is when the resource to be allocated is dollars, 
and $x$ contains the prices of the resource
for each of the agents, denoted $p\in\reals_{++}^m$.
When there are prices, an allocation of $y_i$ dollars provides
$y_i/p_i$ units of some good to agent $i$.
The utility in this case has form
\[
U(x,y;\theta) = \begin{cases}
U(y/p;\theta) & y \geq 0, \quad \ones^Ty \leq B, \\
-\infty & \text{otherwise},
\end{cases}
\]
where the division is meant elementwise,
and $U(z;\theta)$ gives the utility of the agents receiving $z_i$ units
of the resource, $i=1,\ldots,m$.
The resulting convex optimization model $\phi(x;\theta)$ gives the maximum utility
allocation that satisfies the budget constraints, given the current prices.

We can just as well model the allocation of multiple resources,
each with its own budget, across agents or tasks; \eg, we might model the
allocation of computational resources, such as CPU cores, memory, and disk
space, to a pool of tasks. If there are $k$ resources and $m'$ agents, then the
output would be the $k$ allocation vectors for each resource, stacked together
to form a vector $y \in \reals^{m}_{+}$, where $m = km'$.

\paragraph{Utility functions.}
A simple family of utility functions are the separable functions
\[
U(y;\theta) = \sum_{i=1}^m U_i(y_i;\theta),
\]
where $U_i(y_i;\theta)$ is the utility of allocating $y_i$ of the resource
to the $i$th agent or task. In this case the entries of the decision $y$ are
coupled by budget constraints. A simple example for separable utility is
exponential utility $U_i(y_i;\theta) = -\exp(\theta_i y_i) / \theta_i$.

However, $U$ need not be separable.
A common example is when $y$ represents an allocation of a budget
in a portfolio of stocks;
the Markowitz utility or risk-adjusted return is
\[
U(y;\theta) = \mu^Ty - \gamma y^T\Sigma y,
\]
where $\mu\in\reals^m$ is the expected return of each investment, $\Sigma\in\symm_{++}^m$ is
the covariance of the returns,
and $\gamma>0$ is the risk aversion parameter.
We can take $\theta=(\mu,\Sigma,\gamma)$, in which case we are
observing portfolios and attempting to infer the mean
covariance, and risk aversion parameter that best model the observed portfolio
allocations.

\fi

\ifneurips
\else
\section{Stochastic control agent models}\label{s:stoch-control}
\fi
In this section, we consider a setting in which $x \in \inputs = \reals^n$ is
the context or state of a dynamical system, and $y \in \outputs = \reals^m$
represents the action taken by the agent in that state. 
Our goal is to model the agent's actions as coming from a policy,
\ie, a mapping from state to action. In this
section, we describe generic ways to model an agent's policy with a convex
optimization model. The convex optimization models we present are all instances
of convex optimization policies commonly used for stochastic control
\cite{cocp2020}. When learning these models, one can use the same losses
proposed for regression (see \S\ref{sec:regression}).

\ifneurips
\else

\paragraph{Stochastic control.} To motivate the models presented in this
section, we describe here a general stochastic control problem. Let $x_t$ and
$y_t$ denote the state and action at time $t$.
Suppose the state evolves according to the dynamics
\BEQ
x_{t+1} = f(x_t,y_t, w_t),
\label{eq:dynamics}
\EEQ
where $w_t \in \mathcal W$ is a random variable, and the function $f : \reals^{n}
\times \reals^{m} \times \mathcal W \to \reals^{n}$ gives the (stochastic)
dynamics of the dynamical system. Suppose also that the agent selects actions
according to
\BEQ
y_t=\phi(x_t), \quad t=0,1,\ldots,
\label{eq:policy}
\EEQ
where $\phi:\reals^n\to\reals^m$ is the policy, and
that the agent's goal is to minimize a discounted sum of stage costs
$g:\reals^n\times\reals^m\to\reals\cup\{+\infty\}$ over time,
\[
\sum_{t=0}^\infty \gamma^t g(x_t,y_t),
\]
where $\gamma\in(0,1]$ is a discount factor,
subject to the dynamics \eqref{eq:dynamics} and the policy \eqref{eq:policy}.
It is well known (see, \eg, \cite{bertsekas2017dynamic}) that an optimal policy is given by
\[
\phi^\star(x) = \underset{u}{\mbox{argmin}} ~ g(x,y) + \Expect V(f(x,y)),
\]
where $V:\reals^n\to\reals$ is the cost-to-go function,
which satisfies Bellman's equation
\BEQ
V(x) = \underset{u}{\inf} ~ \left( 
g(x,y) + \Expect V(f(x,y))\right), \quad x\in\reals^n.
\label{eq:bellman}
\EEQ

In general, given a dataset describing an agent's actions,
we have no reason to believe that the agent chooses actions by solving
a stochastic control problem. Nonetheless, choosing a model that corresponds to
a policy for stochastic control can work well in practice. As we will see,
our models involve learning the parameters in three functions that can
be interpreted as dynamics, stage costs, and an approximate value function.

\paragraph{Approximate dynamic programming (ADP).}
One possible model of agent behavior is the ADP model
\cite[\S6]{bertsekas2017dynamic},
which has the form
\[
   \begin{array}{llll}
      \phi(x;\theta) &=& \mbox{argmin}_{y} & g(x ,y;\theta) + \hat V(x_+; \theta) \\
     && \mbox{subject to} & x_+ = f(x, y;\theta),
  \end{array}
\]
where $x_{+} \in \reals^{n}$ and $y \in \reals^{m}$ are the variables. The function
$f : \reals^{n} \times \reals^{m} \to \reals^{n}$, which must be affine in its
second argument, can be interpreted as the dynamics;
the function $g : \reals^{n} \times \reals^{m} \to \reals \cup \{+\infty\}$ is the
stage cost (which is convex in its second argument); and the convex function $\hat V$ can
be interpreted as an approximation of the cost-to-go or value function. All
three of these functions are parametrized by the vector $\theta$. The value
$\hat y = \phi(x; \theta)$ is the optimal value of the variable $y$, \ie, the
ADP model chooses the action that minimizes the current stage cost plus an
estimate of the cost-to-go of the next state.

One reasonable parametrized stage cost $g$ is the weighted
sum of a number of convex functions $h_1,\ldots,h_p:\reals^{n} \times \reals^{m}\to\reals\cup\{+\infty\}$,
\ifneurips
$g(x,y;\theta) = \sum_{i=1}^p \theta_i h_i(x,y)$.
\else
\[
g(x,y;\theta) = \sum_{i=1}^p \theta_i h_i(x,y).
\]
\fi
In this case we would have $\Theta=\reals_+^p$.
\ifneurips
\else
For example, if the dynamical system were a car,
the state was the physical state of the car,
and the action was the steering wheel angle and
the acceleration,
there would be many reasonable costs: \eg, tracking,
fuel use, and comfort.
Such a stage cost could be used to trade off these costs,
or to derive them from data.

\fi
Similarly, the cost-to-go function might be a weighted sum of functions $\hat
V_1, \ldots, \hat V_p : \reals^{n} \to \reals \cup \{\infty\}$,
\ifneurips
$\hat V(x;\theta) = \sum_{i=1}^p \theta_i \hat V_i(x);$
\else
\[
\hat V(x;\theta) = \sum_{i=1}^p \theta_i \hat V_i(x);
\]
\fi
\eg, taking
\ifneurips
$\hat V_{i+jn}(x) = x_i x_j$, $i=1,\ldots,n,$  $j=1,\ldots,n$,
\else
\[
\hat V_{i+jn}(x) = x_i x_j, \quad i=1,\ldots,n, \quad j=1,\ldots,n
\]
\fi
yields a quadratic cost-to-go function.

\paragraph{Model predictive control (MPC).}
An MPC policy is an instance of the ADP policy \cite[\S6.4.3]{bertsekas2017dynamic},
\[
 \begin{array}{llll}
   \phi(x;\theta) &=& \mbox{argmin} & \sum_{t=0}^{T-1} g_t(x_t,y_t;\theta) \\
 && \mbox{subject to} & x_{t+1} = f_t(x_t,y_t;\theta), \quad t=0,\ldots,T-1,\\
 &&& x_0 = x,
\end{array}
\]
with variables $x_0,\ldots,x_T$ and $y_0,\ldots,y_{T-1}$,
where $T$ is the time horizon.
Here $f_t:\reals^n\times\reals^m\to\reals^n$ is the (affine) dynamics function
at time $t$, and $g_t:\reals^n\times\reals^m\to\reals\cup\{+\infty\}$ is the stage
cost function at time $t$, which is convex in $y_t$; both functions are
parametrized by $\theta$. (The expression
\ifneurips
$\sum_{t=1}^{T-1} g_t(x_t, y_t; \theta)$
\else
\[
  \sum_{t=1}^{T-1} g_t(x_t, y_t; \theta)
\]
\fi
can be interpreted as the approximate value function of an ADP policy.)
The objective is the sum of the stage costs $g_t$ through time, and
the constraints enforce the dynamics and the initial state.
The MPC model chooses the action as the first action in a planned sequence of
future actions $y_0,\ldots,y_{T-1}$, \ie, $\hat y = \phi(x; \theta)$ is the
optimal value for the variable $y_0$.
\fi

\section{Numerical experiments}\label{s:num-exp}
In this section we present
\ifneurips \else four \fi
numerical experiments that mirror
the examples from sections \S\ref{s:map}
\ifneurips
\else
,\S\ref{sec:utility_max},
\fi
and \S\ref{s:stoch-control}.
The code for all of these examples can be found online at
\[
\verb|https://github.com/cvxgrp/cvxpylayers/tree/master/examples/torch|.
\]

\begin{figure}
\centering
\includegraphics{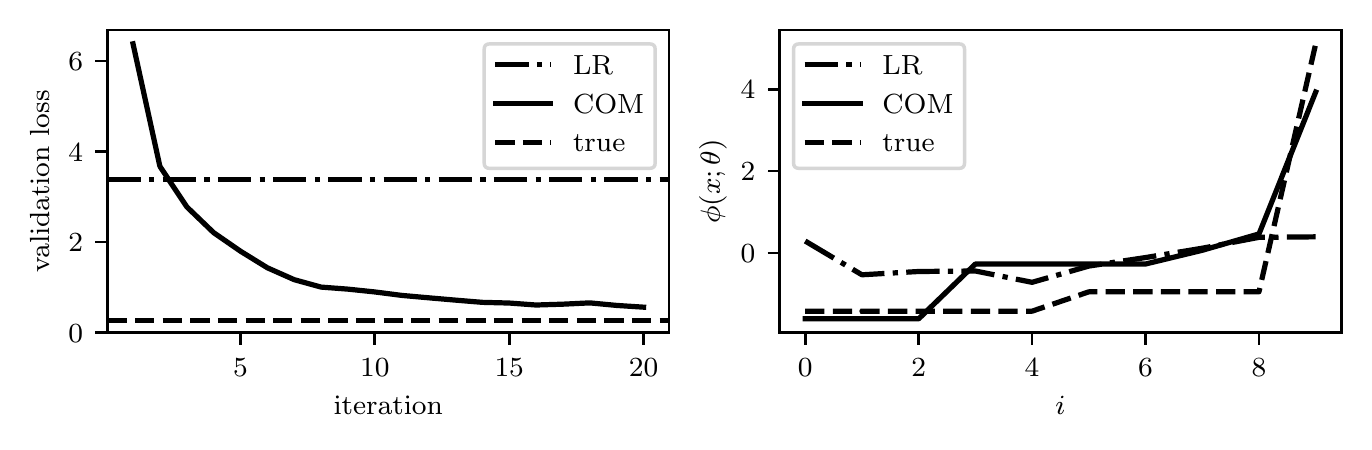}
\caption{Monotonic output regression: linear regression (LR), convex optimization model (COM), true model (true). Left: validation loss.
Right: predictions for a held-out input.}
\label{fig:isotonic_regression}
\end{figure}

\paragraph{Monotonic output regression.}
We consider the monotonic output regression model
(see \S\ref{sec:regression}).
We take $n=20$ and $m=10$.
We generate a true parameter
$\theta^\mathrm{true}\in\reals^{n \times m}$
with entries sampled independently from a standard normal distribution,
and sample 100 training data pairs
and 50 validation data pairs according to
\ifneurips
\else
\[
x \sim \mathcal N(0,I), \quad y = \phi(x + z; \theta^\mathrm{true}),
\quad z\sim\mathcal N(0,I).
\]
\fi
We compare the convex optimization model to linear regression,
using the standard sum of squares loss and no regularizer.
The results for both of these methods
are displayed in figure \ref{fig:isotonic_regression}.
On the left, we show the validation loss
versus training iteration.
The final validation loss for linear regression is 3.375,
for the convex optimization model is 0.562, and for
the true model is 0.264.
We also calculated the validation loss of a convex optimization model
with the linear regression parameters; this resulted
in a validation loss of 1.511.
While better than 3.375, this shows that here our learning method is superior
to learning the parameters using linear regression and then projecting
the outputs onto the monotone cone.
On the right, we show both model's predictions for a validation input.

\paragraph{Signal denoising.}
Here, we fit the parameters in a quadratic MRF (see \S\ref{s:graph}) for
a signal denoising problem.
We consider a denoising problem in which each input $x \in \reals^{n}$
is a noise-corrupted observation of an output $y \in \reals^{m}$, where $n = m$. The goal
is to reconstruct the original signal $y$, given the noise-corrupted
observation. We model
the conditional density of $y$ given $x$ as
\[
  p(y \mid x; \theta) \propto  \exp\left(-\left(\|M(x - y)\|_2^2 + \lambda \|Dy\|_2^2\right)\right),
\]
where $\theta = (M \in \reals^{n \times n}, \lambda \in \reals_{++})$ is the
parameter and $D$ is the first-order difference matrix. The MAP estimate
of $y$ given $x$ is
$\phi(x ; \theta) = \argmax_{y} \log p(y \mid x; \theta)$, and
the corresponding convex optimization model has the energy function
\[
  E(x, y; \theta) = \|M(x - y)\|_2^2 + \lambda \|Dy\|_2^2.
\]
The first term says that $x$ should be close to $y$, as measured by the squared quadratic
$M$-norm, while the second term says that the entries of $y$ should vary smoothly.
When $M = I$, this model is equivalent to least-squares denoising with Laplacian
regularization.
We note that this convex optimization model has the analytical solution
\ifneurips
\else
\[
\phi(x;\theta) = (M^TM+\lambda D^TD)^{-1}M^TM x.
\]
\fi

\begin{figure}
\centering
\includegraphics{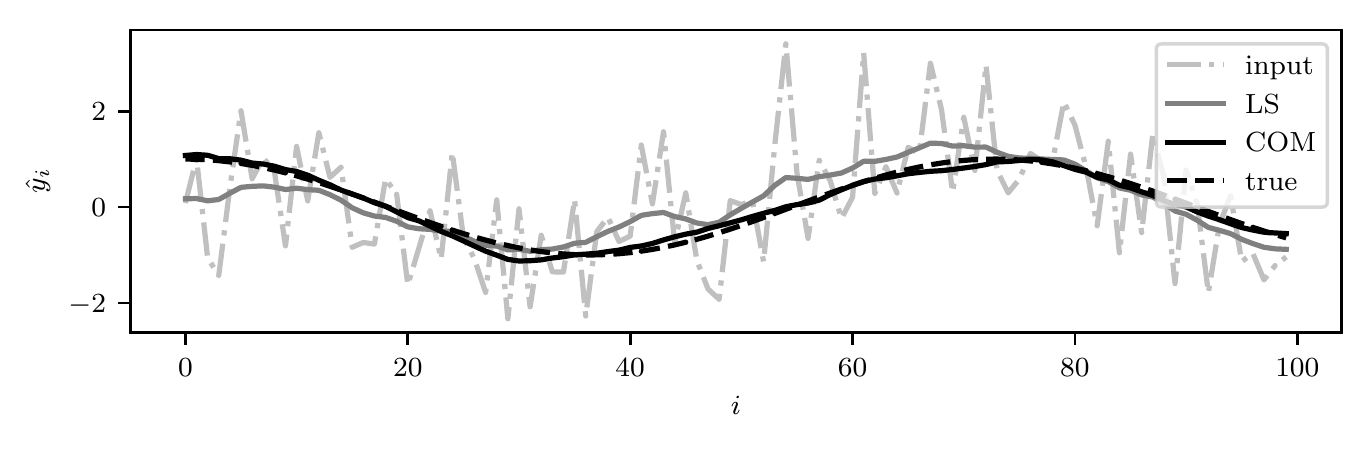}
\caption{Signal denoising. Predictions for a held-out input;
least squares (LS), convex optimization model (COM), and true output (true).}
\label{fig-denoising}
\end{figure}

We use $n = 100$, $m=100$, and $N = 500$ training
pairs. Each output $y$ is generated by sampling a different scale factor $a$
from a uniform distribution over the interval $[1, 3]$, and then evaluating
the cosine function at $100$ linearly spaced points in the interval $[0,
2\pi a]$. The outputs are corrupted by Gaussian noise to produce the inputs.
We generate a covariance matrix
\ifneurips
\else
$\Sigma$ according to
\[
  \Sigma = P^TP, \quad P \sim  \mathcal{N}(0, 0.01I),
\]
\fi
\ifneurips
\else
and then generate the components of each input $x$
\[
  v \sim \mathcal N(0, \Sigma), \quad x = y + v.
\]
\fi
We generate $100$
validation points in the same way. As a baseline, we use least-squares
denoising with Laplacian regularization, sweeping $\lambda$ to find
the value which minimizes the error on the training set. The least-squares
reconstruction achieves a validation loss of 0.090; after learning, the convex
optimization model achieves a validation loss of 0.014.
Figure~\ref{fig-denoising} compares a prediction of the convex optimization
model with least squares and the true output, for a held-out input-output pair.

\ifneurips
\else
\paragraph{Resource allocation.}
We consider an instance of the resource allocation problem
with prices, as described in \S\ref{sec:utility_max},
and use the separable exponential utility function.
The input $x$ consists of the budget $B\in\reals_+$ and the prices $p\in\reals_+^m$,
and the output $y\in\reals_+^m$ is the resource allocation.
Our convex optimization model has the form
\[
 \begin{array}{llll}
  \phi(x;\theta) &=& \mbox{argmin}_y & \sum_{i=1}^m \exp(- \theta_i y_i / p_i) / \theta_i \\
 && \mbox{subject to} & y \geq 0, \quad \ones^Ty \leq B.
\end{array}
\]
Here the feasible parameter set is $\Theta=\reals_+^m$.
We take $m=10$, and sample 100 training and 50 validation inputs and the true parameter according to
\[
B \sim U[0,1], \quad p_i\sim U[0,1], \quad \theta^\mathrm{true}_i \sim U[0,1], \quad i=1,\ldots,m,
\]
where $U[a,b]$ denotes the uniform distribution over the interval $[a,b]$.
The outputs were generated according to
\[
y = B\frac{\phi(B, p; \theta^\mathrm{true}) \odot z}{\ones^T(\phi(B, p; \theta^\mathrm{true}) \odot z)}, \quad z^i\sim U[0.5,1.5], \quad i=1,\ldots,m,
\]
where $\odot$ denotes elementwise multiplication.
In other words, we evaluate the true convex optimization model, multiply each output by a random number between
$0.5$ and $1.5$, and re-scale the allocation so it sums to the budget $B$.
We compare the convex optimization to logistic regression using the prices as features
and the (normalized) allocation as the output.
In figure \ref{fig:resource_allocation} we show results for these two methods.
On the left, we show the validation loss versus iteration for the convex optimization model,
with horizontal lines for the validation loss of the logistic regression baseline and the true model.
On the right, we show the learned and true utility function parameters, and observe that the learned
parameters are quite close to the true parameters.

\begin{figure}
\centering
\includegraphics{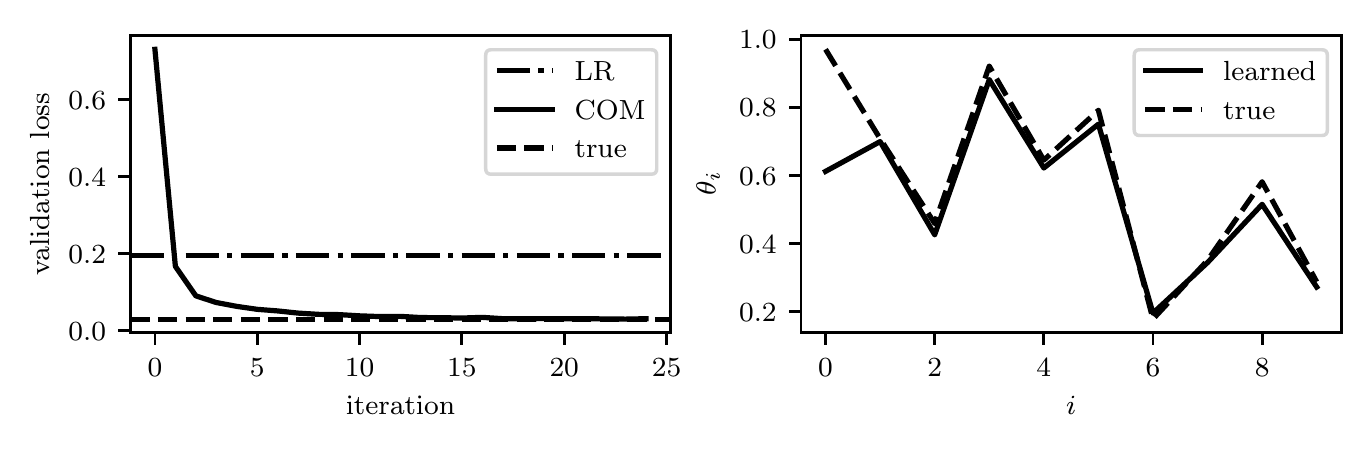}
\caption{Resource allocation. Left: validation loss versus iteration for logistic regression
and our convex optimization model. Right: learned and true parameters.}
\label{fig:resource_allocation}
\end{figure}
\fi

\paragraph{Constrained MPC.}
\begin{figure}
\centering
\includegraphics{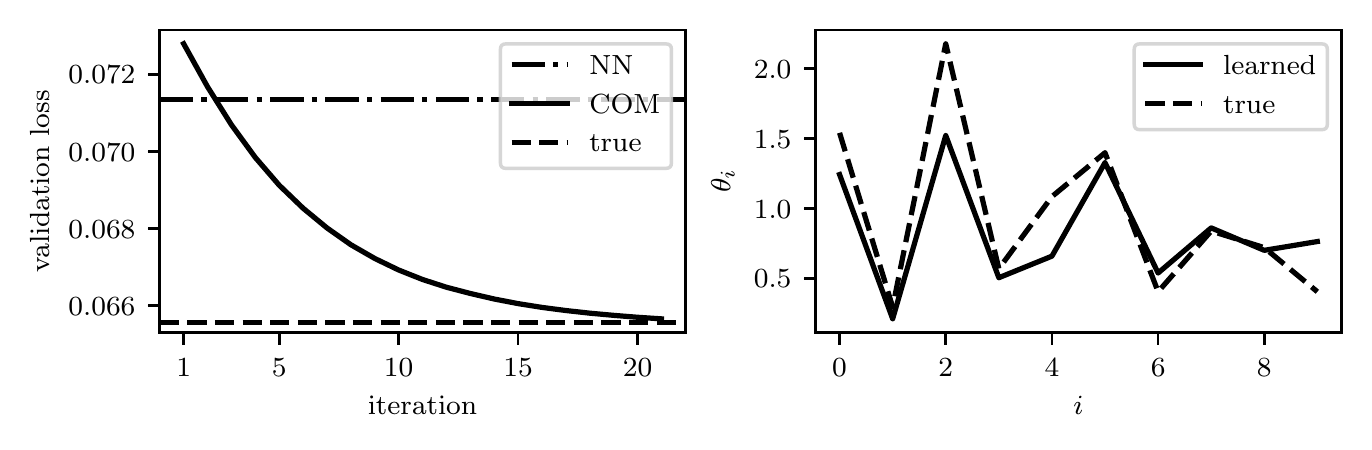}
\caption{Constrained MPC. Left: validation loss for a neural network (NN), convex
optimization model (COM), and true model (true). Right: learned and true
parameters.
}\label{fig:control}
\end{figure}
We fit a convex optimization model for an instance of the MPC problem described
in \S\ref{s:stoch-control}, with $n=10$ states, $m=4$ controls with
$\outputs = \{y \in \reals^{m} \mid \|y\|_{\infty} \leq 0.5\}$, and a horizon
of $T=5$. Our convex optimization model has the form
\begin{equation}
 \begin{array}{llll}
   \phi(x;\theta) &=& \mbox{argmin} & \sum_{t=0}^{T-1} \theta^T x_t^2 + \|y_t\|_2^2 \\
                  && \mbox{subject to} & x_{t+1} = Ax_t + By_t, \quad t=1, \ldots, T-1 \\
                  &&& \|y_t\|_{\infty} \leq 0.5 , \quad t=0, \ldots, T-1\\
                  &&& x_{0} = x,
 \end{array}\label{eq:mpc-policy-instance}
\end{equation}
where the variables are the states $x_0,
\ldots, x_T \in \reals^{n}$ and the controls $y_0, \ldots, y_{T-1} \in
\reals^{m}$, the square $(x_t)^2$ is meant elementwise, and $\phi(x; \theta)$
is the optimal value of $y_0$. The parameter $\theta \in \reals^{n}_{+}$
parametrizes the stage cost, and the dynamics matrices $A$ and $B$ are known
numerical constants.

We generate
a true weight $\theta^\mathrm{true} \in \reals^{n}_{+}$ with entries set to the
absolute value of samples from a standard normal distribution. The dataset is
generated by rolling out an MPC policy $\phi^\mathrm{true}$ of the form
(\ref{eq:mpc-policy-instance}), with parameter $\theta^{\mathrm{true}}$. The
policy is simulated from an
initial state $x_0 \sim \mathcal{N}(0, I)$. The outputs are noise-corrupted
controls, generated according to
\[
  u^i = \phi^{\mathrm{true}}(x^i), \,\,
  z^i \sim \mathcal{N}(0, 0.1I), \,\,
  y^i = \Pi_{\outputs}(u^i + z^i), \,\,
  \nu^{i-1} \sim \mathcal{N}(0, I), \,\,
  x^i = Ax^{i-1} + B u^{i-1} + \nu^{i-1},
\]
for $i = 1, \ldots, 1000$. We generate $1000$ validation points in the same way.

We use the mean-squared loss for the loss function $L$, and train for
20 iterations. As a baseline, we compare against a two-layer feedforward ReLU
network with hidden layer dimension $n$, and with output clamped to have absolute
value no greater than $0.5$.
The results are displayed in figure~\ref{fig:control}.
The ReLU network achieves a validation loss of 0.071. The trained convex
optimization model achieves a validation loss of 0.066, which is close to the
validation loss of the underlying model. Additionally, the convex optimization
model nearly recovers the true weights.

\clearpage

\ifneurips
\else
\section*{Acknowledgements}
Akshay Agrawal is supported by a Stanford Graduate Fellowship.
Shane Barratt is supported by the National Science Foundation Graduate Research Fellowship
under Grant No. DGE-1656518.
\fi

\printbibliography{}

\end{document}